\documentclass{article}
\usepackage{amsmath,amssymb,amstext,amsfonts}
\usepackage{graphicx}
\usepackage{algorithm}
\usepackage{algpseudocode}
\usepackage{longtable}
\usepackage{booktabs}
\usepackage{multirow}

\usepackage{color}
\usepackage{soul}
\usepackage{ulem}
\usepackage{subfig}

\providecommand{\keywords}[1]{\textbf{\textit{Index terms---}} #1}

\begin{document}

\title{Short-Term Temporal Convolutional Networks for Dynamic Hand Gesture Recognition}

\author{Yi Zhang\thanks{Faculty of Electrical Engineering and Computer Science, Ningbo University, P. R. China}, Chong Wang\thanks{Corresponding author (e-mail: wangchong@nbu.edu.cn)}, Ye Zheng, Jieyu Zhao, Yuqi Li and Xijiong Xie
\thanks{This work was supported in part by the National Natural Science Foundation of China (61603202, 61571247), Zhejiang Provincial Natural Science Foundation of China (LY20F030005) and K.C. Wong Magna Fund in Ningbo University.}
}
\maketitle

\begin{abstract}
The purpose of gesture recognition is to recognize meaningful movements of human bodies, and gesture recognition is an important issue in computer vision.
In this paper, we present a multimodal gesture recognition method based on 3D densely convolutional networks (3D-DenseNets)  and improved temporal convolutional networks (TCNs).
The key idea of our approach is to find a compact and effective representation of spatial and temporal features, which orderly and separately divide task of gesture video analysis into two parts: spatial analysis and temporal analysis. 
In spatial analysis, we adopt 3D-DenseNets to learn short-term spatio-temporal features effectively.
Subsequently, in temporal analysis, we use TCNs to extract temporal features and employ improved Squeeze-and-Excitation Networks (SENets) to strengthen the representational power of temporal features from each TCNs' layers.
The method has been evaluated on the VIVA and the NVIDIA Gesture Dynamic Hand Gesture Datasets. 
Our approach obtains very competitive performance on VIVA benchmarks with the classification accuracies of $91.54\%$, and achieve state-of-the art performance with $86.37\%$ accuracy on NVIDIA benchmark.

\end{abstract}

\keywords{Gesture Recognition,  3D-DenseNets,  TCNs,  multimodal.}

%

\section{Introduction}
\label{sec:introduction}
\begin{figure}[t]
\begin{center}
\includegraphics[width=0.45\textwidth]{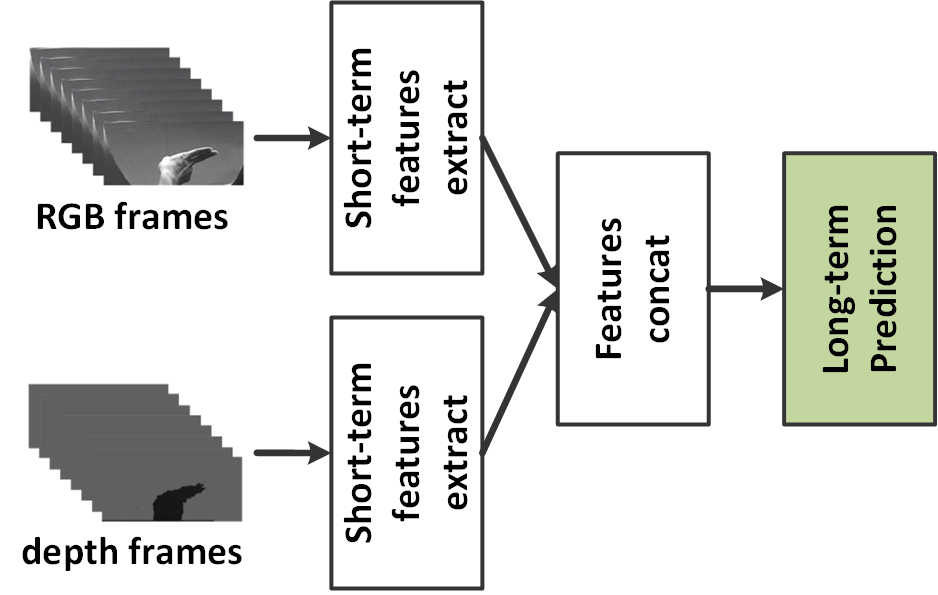}
\caption{An overview of the proposed method. The proposed deep architecture is composed of two main steps: (a) Multimodal short-term spatio-temporal feature sequence extracting by truncated 3D-Densenet (T3D-Dense), local temporal average pooling (LTAP) and multimodal features concatenation. (b)Long-term feature sequence recognizing via TCN and TSE.}
\label{fig:pipline}
\end{center}
\end{figure}

Gesture recognition is a fast expanding field with applications in human-computer interaction\cite{rautaray2015vision}, sign language recognition\cite{camgoz2017subunets} and etc.. 
Due to subtle differences among similar gestures, complex scene background, different observation conditions, and noises in acquisition, robust gesture recognition is very challenging.\\
\indent
The main task of gesture recognition is to extract features from an image or a video and then classify or determine each sample to a certain label. 
Gesture recognition aims to recognize and understand meaningful movement of human bodies in which arms and hands play crucial roles. 
Only few gestures can be identified from their spatial or structure information in an image or a single frame. 
In fact, motion cues and structure information simultaneously characterize a unique gesture. 
How to learn spatiotemporal features effectively is always the key in gesture recognition. 
Although in the past decades, many methods have been proposed for this issue, ranging from static to dynamic gestures, and from motion silhouettes-based to the convolutional neural network-based, there are still many challenges associated with the recognition accuracy.\\
\indent
At present, although most existing models have reached a high performance for isolated gesture recognition, most methods have been developed based on Convolutional Neural Networks (CNNs)\cite{simonyan2014two}\cite{neverova2014multi} or Recurrent Neural Networks (RNNs)\cite{molchanov2016online}. With the development of deep learning, more and more new architectures of CNNs have been proposed, especially DenseNets \cite{huang2017densely} what have powerful feature extraction ability. Meanwhile, a new architecture to solve sequence problem named TCNs \cite{bai2018empirical} have been proposed. Compared to RNNs and their canonical recurrent architectures such as LSTMs and GRUs, TCNs have comparable clarity and simplicty. In our approach, we adopt 3D-DenseNet to extract short-term stapio-temporal features, then these features are input into the TCNs to finish the task of Classification.\\
\indent
However, recently, for extracting more complete temporal features, a few methods have been proposed based on attention mechanism. The research prove that there are various relationships between features' interior in neural networks. SENets \cite{hu2018squeeze} are new architectural unit with the goal of improving the quality of representations produced by a network by explicitly modelling the interdependencies between the channels of its convolutional features. And in our approach, we reform SENets and combine them into TCNs to strengthen capacity of TCNs in temporal features extracting.\\
\indent
The pipline of our method is depicted in Figure \ref{fig:pipline}, and the main contribution can be summarized as following:
\begin{itemize}
  \item \textbf{Spatial analysis.} We design a multi-stream truncated 3D-DenseNet, which extracts spatio-temporal features from a video, and through local temporal pooling, obtain the decomposed short-term spatio-temporal features, to solve problem that single frame image can not carry enough spatial or structure information of gesture and reduce repetitive training for video clips.
  \item \textbf{Temporal analysis.} We employ TCN to replace RNN as the main model of sequence information feature analysis. In addition, we improve SENets and apply them in temporal domain to rescale the weights between temporal features and extract more effective temporal features to achieve higher classification accuracy.
\end{itemize}

\section{Related Work}
\label{sec:relatedwork}
Gesture taxonomies and representations have been studied for decades.The vision based gesture recognition techniques include the static gesture oriented and the dynamic gesture oriented methods \cite{rautaray2015vision}.\\
\indent
Recently, convolution neural networks (CNNs) \cite{krizhevsky2012imagenet} have made a great breakthrough on computer vision related tasks by their powerful feature extraction ability, thus the features extracted by CNNs are widely used in many action classification tasks instead of hand-crafted features for better performance. 
Features are extracted by 2D-CNN from the starts. 
bi-directional rank pooling \cite{wang2016large}\cite{fernando2016rank} was used to encode the spatial and temporal information of videos.
Temporal convolutions for gesture recognition in videos
Beyond temporal pooling \cite{pigou2018beyond}  was proposed to solve gesture recognition problem in videos by a new temporal pooling method.
On the other hand, C3D\cite{tran2015learning} model is developed and provides a better performance and main contribution in this research is proposed an architecture to extract spatio-temporal features from a video clip. Concurrently, a multi-stream 3D-CNN\cite{molchanov2015hand} was designed for hand gesture recognition and the classifier consisted of two subnetworks: a high-resolution network (HRN) and a low-resolution network (LRN) in this model.\\
\indent
Meanwhile, with the development of convolutional neural networks, more and more architectures of CNNs were proposed, like AlexNet \cite{krizhevsky2012imagenet}, VGGNet \cite{simonyan2014very}, GoogleNet \cite{szegedy2015going} \cite{ioffe2015batch} \cite{szegedy2016rethinking} \cite{szegedy2017inception}, ResNet \cite{he2016deep} and DenseNet \cite{huang2017densely}. All of these models have one target that is building a higher architectures of CNNs to dig deeper and more complete statial features from low-level image frames, and then classify. 
In the area of isolated gesture recognition, Res-C3D model\cite{miao2017multimodal} was used and won the first place twice in ChaLearn LAP Multi-modal Isolated Gesture Recognition Challenges 2016 \cite{escalante2016chalearn} and 2017 \cite{wan2017results}. Whatmore, DenseNets as one of the latest convolutional architectures, was adopted in action recognitions especially face recognitions and gesture recognitions gradually. A face recognition model named Dense Face\cite{zhang2018face} was proposed to explore the performance of densely connected network in face recognition. 
DenseNets\cite{hao2019spatiotemporal} also was used to classifier the different actions in recent researches.
\\
\indent
Regarding the temporal information of the video sequences, Long Short Term Memory(LSTM) networks is a common choice to gesture recognition. 
For instance, convolutional LSTM\cite{zhang2017learning} was introduced for spatio-temporal feature maps. 2S-RNN(RGB and Depth)\cite{chai2016two} was used for continuous gesture recognition.
However, RNNs including LSTMs and GRUs have some weaknesses on temporal domain like short-range information learning, oversized memory capacity.
To make these weaknesses up, TCNs is proposed and applied in the gesture reconition.
Res-TCN\cite{hou2018spatial} was proposed  for skeleton-based dynamic hand gesture recognition.
Whatmore, a model based on TCN\cite{tsinganos2019improved} was proposed for gesture recognition.\\
\indent
Other important works based on attention mechanism. Attention mechanism or attention model firstly was applied to neural networks by Vaswani et al\cite{vaswani2017attention}. After that, more and more researches are proposed based on attention mechanism, so as SENets \cite{hu2018squeeze} that improve ResNets to win first place of ILSVRC 2017 classification.

\begin{figure*}[tp]
\begin{center}
\includegraphics[width=0.95\textwidth]{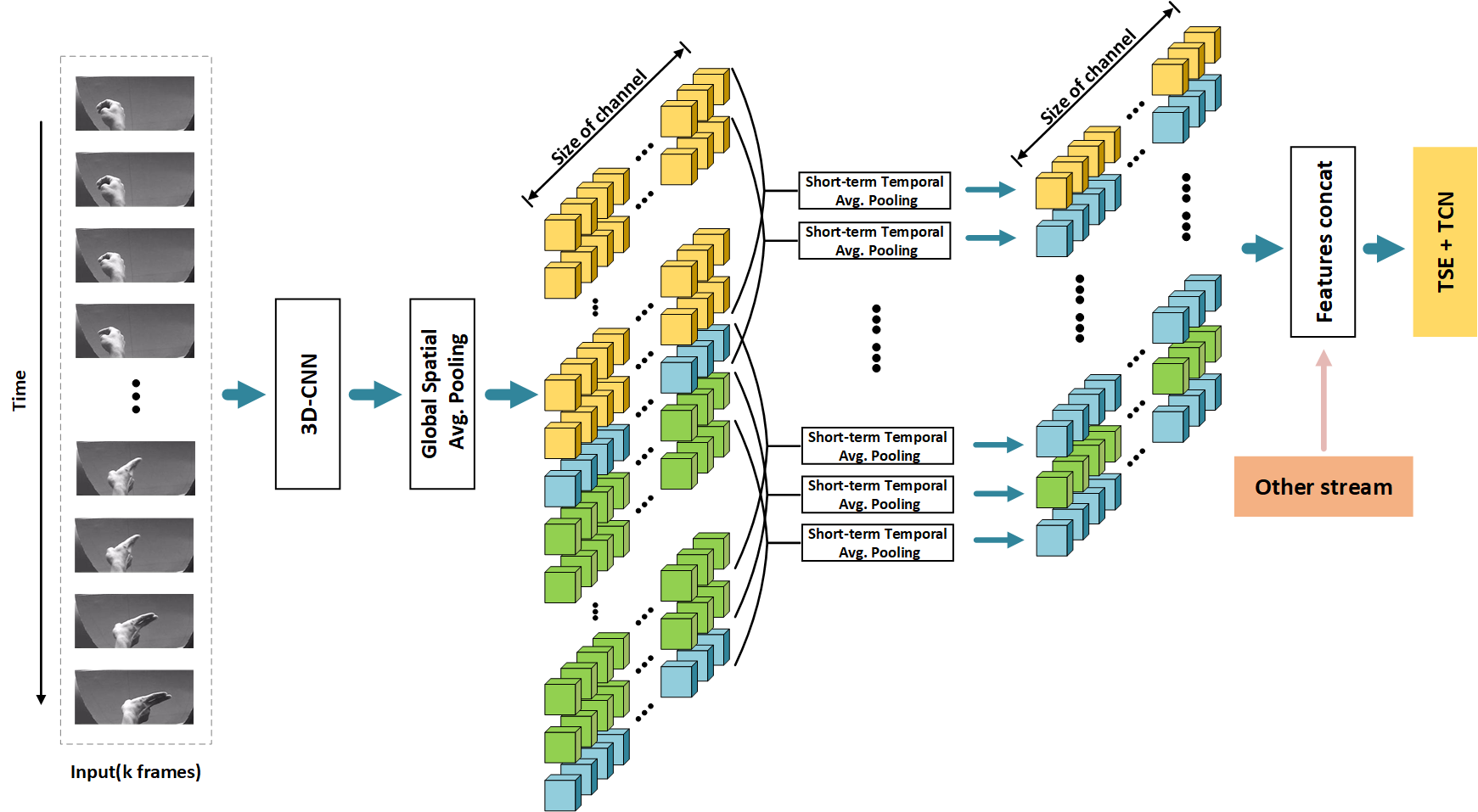}
\label{fig:dense}
\caption{The architecture of 3D-DenseNet.}
\end{center}
\end{figure*}

\section{Our Approach}
\label{sec:approach}
In the video recognition, both of the spatial and temporal information are important. Although there have been impressive progress in spatial feature extraction using 2D-CNNs based networks\cite{molchanov2015hand}\cite{simonyan2014two}, how to effectively learn the temporal features is still a very challenging problem. Unlike the 2D-CNNs focusing on the single image, various 3D-CNN based networks\cite{molchanov2016online}\cite{wang2016robust}\cite{carreira2017quo}\cite{abavisani2019improving} have been proposed to process the successive frames simultaneously. For the video of dynamic hand gestures, adjacent frames are usually similar and containing the same static gesture, while the static gestures change several times during the whole video. Thus, in this paper we decompose the video to two different parts. One is the short-term spatio-temporal information in the adjacent frames, and the other is the long-term temporal information analysed by a sequential model. Based on this consideration, we raised two major questions,
\begin{itemize}
\item how to learn short-term spatio-temporal features effectively from video clips in the same video. 
\item how to reasonably classify a sequence which is combined from these consecutive features. 
\end{itemize}

In order to address these issues, we designed a novel architecture to extract a sequence of short spatio-temporal features in order to recognize dynamic gestures. 

\begin{figure}[t]
\begin{center}
\includegraphics[width=0.48\textwidth]{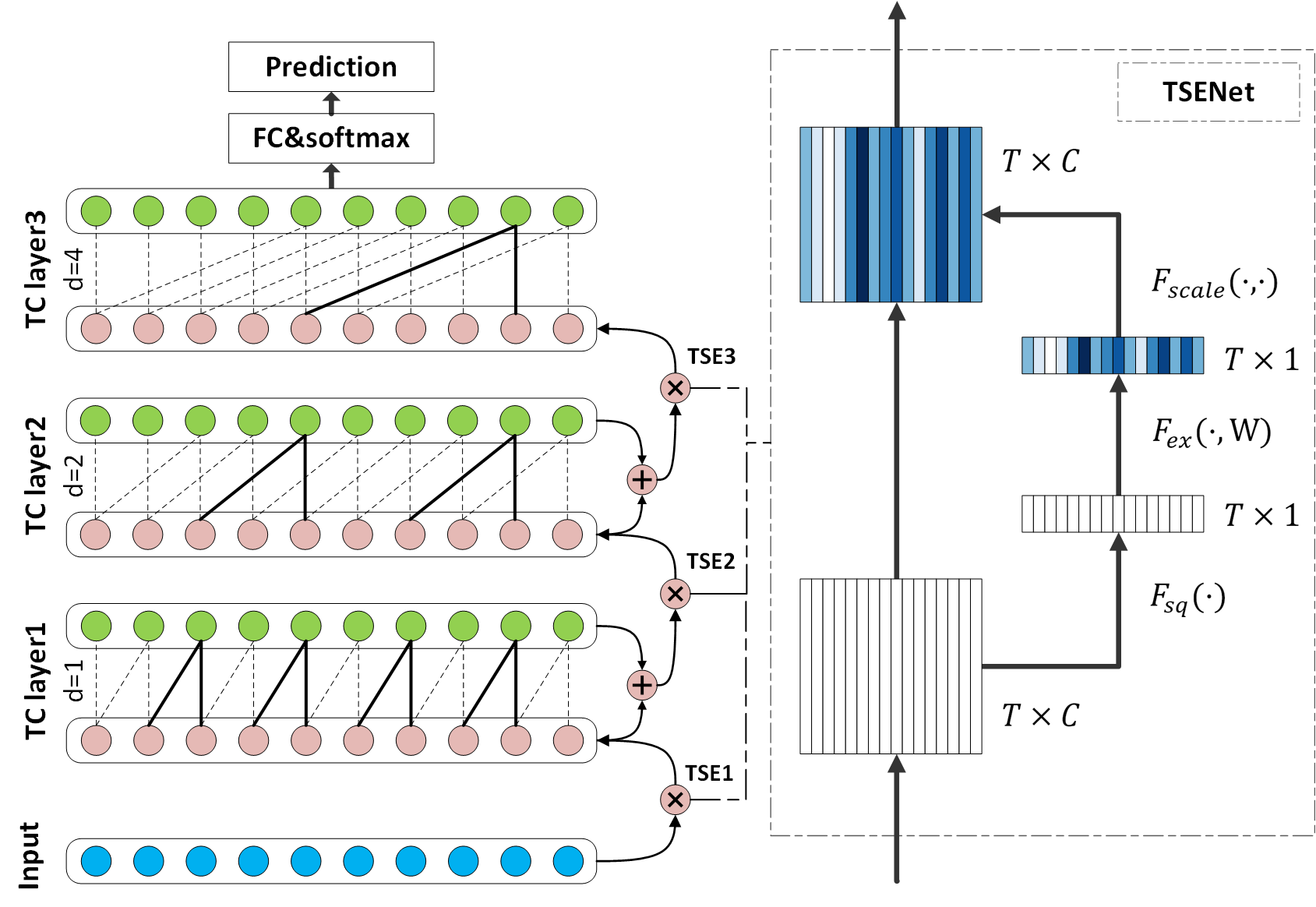}
\label{fig:TCN}
\caption{Architecture and architectural elements in a TCN. There is an example that dilated causal convolution with dilation factors d = 1,2,4 and filter size k = 2 in figure. The receptive field is able to cover all values from the input sequence. And adjacent layers are connected by residual block.Before temporal convolution layer, the inputs need to go through the corresponding Temporal Squeeze-and-Excitation(TSE) layer to adjust weight of input in temporal domain.}
\end{center}
\end{figure}

As depicted in Figure \ref{fig:pipline}, the overall process can be divided into two parts: 1) multi-modal short-term spatio-temporal feature extraction based on 3D-DenseNets and 2) spatio-temporal sequence classify with and temporal SENets embedded TCNs. To be specific, the details of the proposed network structure is presented in Figure \ref{fig:dense} and Figure \ref{fig:TCN}.

\subsection{temporal local pooling to extract short-term features}

\indent
Due to the availability of various data types and the nature of signing videos, a more robust feature representation can acquired from the incorporation of multi-modal hand gesture information. To effectively present the the location, shape and sequential information in the adjacent gesture frames, we design a multi-stream DenseNet based on the C3D\cite{tran2015learning} to extracts short-term spatio-temporal features.
Assume a given video $V$ with $n$ frames, it is firstly re-sampled to $k$ frames. Thus, the input video $V_{S}$ is denoted as,
\begin{equation}V_{S}=[v_{1},v_{2},...,v_{k}]\label{eq}\end{equation}
where $v_{k}$ is the $k$-th frame image of video sequence in the input.\\
\indent
As aforementioned, we consider multiple modalities of gesture video data as the input. Each type of the data is set as one data stream and fed to the same network structure. The outputs of them will be fused together later as shown in Figure \ref{fig:pipline}.
The proposed model contains 4 dense blocks, containing 6, 12, 24, 16 layers respectively. Following the basic design in DenseNet\cite{huang2017densely}  and C3D\cite{tran2015learning}, the detailed network configurations are shown in Table 1. It is worth noting that most of the convolution layers are with $3\times3\times3$ filters, which limits the process only on the local spatial and temporal domain. Moreover, the temporal pooling size and stride in all the transition layers are set as 1 to avoid the fusion of the short-term temporal information, which is one major difference from the other conventional 3D-CNNs\cite{tran2015learning}.

\begin{table}
\centering
\caption{\textbf{3D-DenseNet architectures. The growth rate of network is k = 12. Note that each ``conv'' layer shown in the corresponds the sequence BN-ReLU-Conv.}}
\label{table:dense}
\setlength{\tabcolsep}{3pt}
\begin{tabular}{cc}

\toprule
Layers 
 & Filter Size\\  
 \midrule

  Convolution 
  & $5\times5\times5$ conv, stride $2\times2\times1$ \\  
  
  Pooling 
  & $3\times3\times1$ max pool, stride $2\times2\times1$ \\    
  
Dense Block 1 
&
$\begin{bmatrix}
1 \times 1 \times 1\; conv\\ 
3 \times 3 \times 3\; conv
\end{bmatrix} \times 6 $  \\


\multirow{2}{*}{Transition Layer 1} 
& $1\times1\times1$ conv\\
									& $2\times2\times1$ average pool, stride $2\times2\times1$\\

Dense Block 2 
&
$\begin{bmatrix}
1 \times 1 \times 1\; conv\\ 
3 \times 3 \times 3\; conv
\end{bmatrix} \times 12 $  \\

\multirow{2}*{Transition Layer 2} 
& $1\times1\times1$ conv\\
& $2\times2\times1$ average pool, stride $2\times2\times1$\\

Dense Block 3 
&
$\begin{bmatrix}
1 \times 1 \times 1\; conv\\ 
3 \times 3 \times 3\; conv
\end{bmatrix} \times 24 $  \\
\multirow{2}*{Transition Layer 3} 
& $1\times1\times1$ conv\\
& $2\times2\times1$ average pool, stride $2\times2\times1$\\

Dense Block final 
&$\begin{bmatrix}
1 \times 1 \times 1\; conv\\ 
3 \times 3 \times 3\; conv
\end{bmatrix} \times 16 $  \\
\multirow{3}*{Classification Layer} 
&  global spatial average pool\\
&  global temporal average pool\\
&fully-connected, softmax\\

 \bottomrule

\end{tabular}
\label{tab1}
\end{table}

Since the 3D-Densenet is served as a short-term spatio-temporal features extractor, we truncate it to obtain the features only. To be specific, the global temporal average pooling layer, last softmax and fully-connected layers are discarded, after the model is first pre-trained with isolated gesture data.

Therefore, we can get the global spatio-temporal feature $F_{k}$ after the global spatial average pool layer,
\begin{equation}F_{k}=[f_{1},f_{2},…,f_{k}]\end{equation}
where the temporal length is $k$ and represent respective spatial feature of $k$ frames. 

Then $T$ short-term spatio-temporal features are cut and pooled from the global feature $F_{k}$. The $t$-th short-term spatio-temporal feature $x_{t}$ is constructed as,
\begin{equation}x_{t}=ltap[f_{t-\frac{k}{T}},f_{t-\frac{k}{T}+1},…,f_{t+\frac{k}{T}-1}]\end{equation}
where $ltap$ is local temporal average pool layer in truncated 3D-Densenet, $\frac{k}{T}$ is half of temporal feature interval. In this way, the adjacent $ltap$ windows also overlapping that assure the relevance and completeness of the front and back frame information.\\
\indent
After local temporal average pooling, we can get a sequence of short-term features in single modality. Multimodal feature sequences are fused into one sequence before input into TCN. In this paper, all feature sequences of different modality are concated in channel dimension.

\begin{figure*}[tp]
\begin{center}
\includegraphics[width=0.96\textwidth]{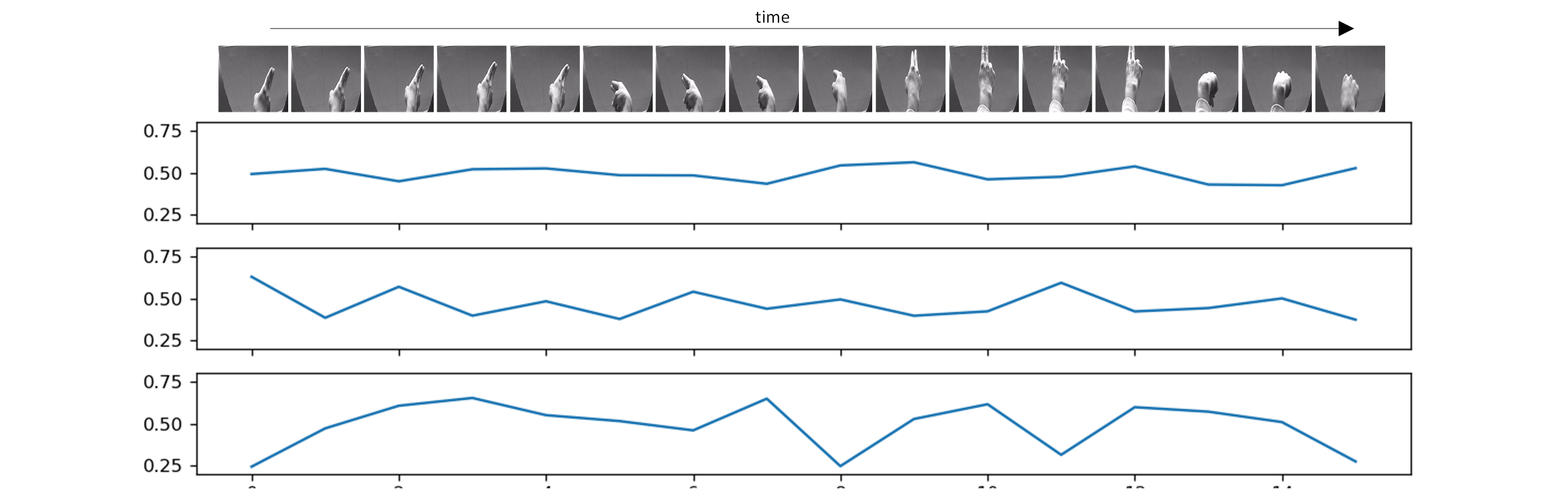}
\label{fig:instance}
\caption{An example sequence from VIVA gesture and its corresponding temporal weghts from TSE-Nets.}
\end{center}
\end{figure*}


\subsection{TSENet + TCN for long-term prediction}
\indent
Based on the short-term spatio-temporal features extracted from all kinds of data modalities (RGB, optic flow, depth, etc.), the long-term temporal features of the whole video is considered to classify the category of the given hand gesture. In this work, a sequence recognition model named TCNs is employed and modified to process the long-term temporal information. The main characteristics of TCNs are the use of causal convolutions and the mapping of an input sequence to an output sequence of the same length. In addition, accounting for sequences with long history, this model uses dilated convolutions that enable a large receptive field as well as residual connections that allow training deeper networks. Considering that our task is to classify the category of hand gesture videos, the output layer of TCN is further processed by one fully connection layer to obtain a single class label for each gesture sequence. The structure of the proposed modified version of the TCN model is depicted in Figure \ref{fig:TCN}.

The short-term temporal features $X=[x_{1},…,x_{T}]$ are utilized as the input sequence of the proposed modified TCN with the outputs $Y=[y_{1},…,y_{T}]$, while the calculation of $y_{t}$, $t<T$ depends only on {$X=[x_{1},…,x_{T}]$}. The reason is that the dilated convolutions are calculated as,
\begin{equation}y_{t}=(x*_{d}h)_{t}=\sum x_{t-d_{m}}h_{m}\end{equation}
where $*_{d}$ is the operator for dilated convolutions, $d$ is the dilation factor and $h$ is the filter's impulse response. For a TCN with $L$ layers, the output of the last layer $y^L$ is used for the sequence classification. 
The class label $\widehat{o}$ attributed to the sequence is found through a fully connected layer with a softmax activation function,
\begin{equation}\widehat{o}=softmax(W_{o}\cdot y^L + b_{o})
\label{eqn::softmax}
\end{equation}
where $W_{o}, b_{o}$ are trainable parameters. \\
\indent
It is noting that the short-term spatio-temporal features $x_{1},…,x_{T}$ actually have different contributions to the recognition in the long-term temporal information processing. For instance, the gesture "swipe +" in Figure 4(b) contains three paths. The first path is extremely similar to the gesture "swipe left" (Figure 4(a)) when $t<9$. The same phenomenon occurs between the third path of the gestures "swipe +" and "swipe down" (Figure 4(c)) when $t>23$. In order to assign different temporal weight to $X=[x_{1},…,x_{T}]$ ,  a temporal Squeeze-and-Excitation network (TSENet) block is inserted between each temporal convolution layers. 

As shown in Figure \ref{fig:TCN}, the average pooling is applied on the channel dimensions $C$ of  $X=[x_{1},…,x_{T}]$ to squeeze channel-wise information. Such obtained temporal descriptor $z=[z_{1},…,z_{T}]$ is a $T\times1$ vector, while the $t$-th element of $z$ is calculated as,
\begin{equation}z_{t}=F_{sq}(x_{t})=\frac{1}{C}\sum_{i=1}^{C}x_{t}(i)\end{equation}

Then another excitation operation is followed to capture the temporal dependencies, i.e. the temporal weights. To fulfil this objective, we opt to employ a simple gating mechanism with the activations:

\begin{equation}s=F_{ex}(z,W)=\sigma (g(z,W))=\sigma (W_{2}\delta (W_{1}z)))\end{equation}
where $\sigma$ refers to the sigmoid fuction, $\delta$ refers to the ReLU fuction, $W_{1}\in \mathbb{R}^{\frac{T}{r}\times T}$ and $W_{2}\in \mathbb{R}^{T\times \frac{T}{r}}$, and $r$ is the size of squeeze channel.The final output of the block is obtained by rescaling the transformation output $U$ with the activations:
$$\widetilde{x}=F_{scale}(u_{t},s_{t})=s_{t}\cdot u_{t}$$
where $\widetilde{X}=[\widetilde{x}_{1},\widetilde{x}_{2},...,\widetilde{x}_{T}]$ and $F_{scale}(u_{t},s_{t})$ refers to temporal-wise multiplication between the scalar $s_{t}$ and the feature map $u_{t}\in \mathbb{R}^{T}$.

An example of  the weights on different TSENet layers is illustrated in Figure 5. It can be seen that the values of the weights changes corresponding to the input gesture sequence as desired.

\section{Experiment}
\indent
The proposed network architecture is implemented by tensorflow, and trained using one NVIDIA Quadro GP100 GPU.  Multimodal 3D-DenseNet models have same structures and are pre-trained using RGB and optic flow(if optic flow existed or can be calculated) data respectively. Adam optimizer is used for training 3D-DenseNet and the learning rate is initialized to $6.4e-4$ and decayed by 10 every 25 epochs. The weight decay is set to $1e-4$. And the dropout rate is set to 0.2. The compression rate and the growth $k$ in the DenseNet block are set as 0.5 and 12, respectively. For the TCN model, we use Adam optimizer for training, and the learning rate is initialized to $1e-4$, epsilon is $1e-8$. 

\subsection{Dataset}
\indent In this section,  we compare our method with the other state-of-the-art dynamic hand gesture methods. Two publicly available multi-modal dynamic hand gesture datasets (VIVA\cite{molchanov2015hand} and NVGesture\cite{molchanov2016online}) are used to evaluate our proposed model in the experiments.

\textbf{VIVA}\cite{molchanov2015hand} 
The VIVA challenge’s dataset is a multimodal dynamic hand gesture dataset specifically designed with difficult settings of cluttered background, volatile illumination, and frequent occlusion for studying natural human activities in real-world driving settings. This dataset was captured using a Microsoft Kinect device, and contains 885 intensity and depth video sequences of 19 different dynamic hand gestures performed by 8 subjects inside a vehicle. Figure 4 shows some gesture sequences.

\textbf{NVGesture}\cite{molchanov2016online}
The NVGesture dataset has been captured with multiple sensors and from multiple viewpoints for studying human-computer interfaces. It contains 1532 dynamic hand gestures recorded from 20 subjects inside a car simulator with artificial lighting conditions. This dataset includes 25 classes of hand gestures. The gestures were recorded with SoftKinetic DS325 device as the RGB-D sensor and DUO-3D for the infrared streams. In the experiments, we use RGB, depth and optical flow modalities, while the optical flow is calculated from the RGB stream using the method presented in \cite{farneback2003two}.

\subsection{Data Preprocessing}

In VIVA dataset, data augmentation is comprised of three other operations: reverse ordering of frames, horizontal mirroring, and applying both operations together. With these operations we generated additional samples for training. For example, applying both operations transforms the original gesture "Swipe Left" with the right hand to a new gesture "Swipe Left"  with the left hand . In NVGesture dataset, for special augmentation, videos are resized to have the smaller video size of 256 pixels, and then randomly cropped with a 224x224 patch.

Data normalization is also applied on both datasets, since a fixed dimension of input data is required in the C3D model and TCN model. For the videos with different temporal lengths, uniform normalization with temporal upsampling and downsampling is used. To compress or extend a given video $V$ with $n$ frames to $k$ frames, 1) If $n>k$, we split the video $V$ into a $k$ section video set $V_{S}$ averagely, where $V_{S}=[V_{1},V_{2},...,V_{k}]$. For each piece in the video set $V_{S}$, we randomly choose one frame as the representation of the sub-video fragment. Finally we concatenate all the represent frames and make them as the result of the normalization. 2) If $n<k$, we randomly choose $k-n$ frame in the video, then repeat them follow by themselves. 

In our experiments, the average number of frames $k$ is set as 32 for VIVA dataset and 64 for NVGesture dataset. Due to the high complexity of 3D convolutional calculating, the spatial size of the inputs is restricted to $112\times112$.\\

\begin{table}[b]
\caption{\textbf{Accuracies of different multimodal fusion-based hand gesture methods on the VIVA dataset. The top performer is denoted by boldface.}}
\label{table:VIVA}
\centering
\setlength{\tabcolsep}{3pt}
\begin{tabular}{lcc}
\toprule
Method & Fused modalities &Accuracy\\  
  \midrule
HOG+HOG2\cite{ohn2014hand}&RGB+Depth&64.5 \\
CNN:LRN\cite{molchanov2015hand}&RGB+Depth&74.4 \\
CNN:LRN:HRN\cite{molchanov2015hand}&RGB+Depth&77.5\\
C3D\cite{tran2015learning} &RGB+Depth&77.4 \\
I3D\cite{carreira2017quo}  &RGB+Depth&83.10 \\
MTUT\cite{abavisani2019improving} &RGB+Depth&86.08 \\
3D-Dense&RGB+Depth&88.21\\
Res3D+TCN &RGB+Depth&85.97 \\
3D-Dense+TCN &RGB+Depth&90.73 \\
3D-Dense+TCN$_{tse}(proposed)$&RGB+Depth&\textbf{91.54}\\

  \bottomrule

\end{tabular}
\label{tab1}
\end{table}

\subsection{Evaluation on VIVA Dataset.}
\indent
Table \ref{table:VIVA} shows the performance of the dynamic hand gestures tested on the RGB and depth modalities of the VIVA dataset. The compared methods include the hand-crafted approach HOG+HOG2\cite{ohn2014hand}, the recurrent CNN-based method(CNN:LRN)\cite{molchanov2015hand}, the C3D model which were pretrained on Sport-1M dataset, the I3D method\cite{carreira2017quo} that performs very well in action recognition, and the Multimodal Training / Unimodal Testing (MTUT) model\cite{abavisani2019improving} which shows promising performance in dynamic hand gesture recognition. All the results are reported by averaging the classification accuracies. It can be seen that the proposed model achieves the highest accuracy, which is 5.46\% higher than the state-of-the-art method MTUT.
This experiment shows that our model is effective to extract both short-term and long-term spatio-temporal information for dynamic hand gesture recognition.

\begin{figure}[t]
\begin{center}
\includegraphics[width=0.45\textwidth]{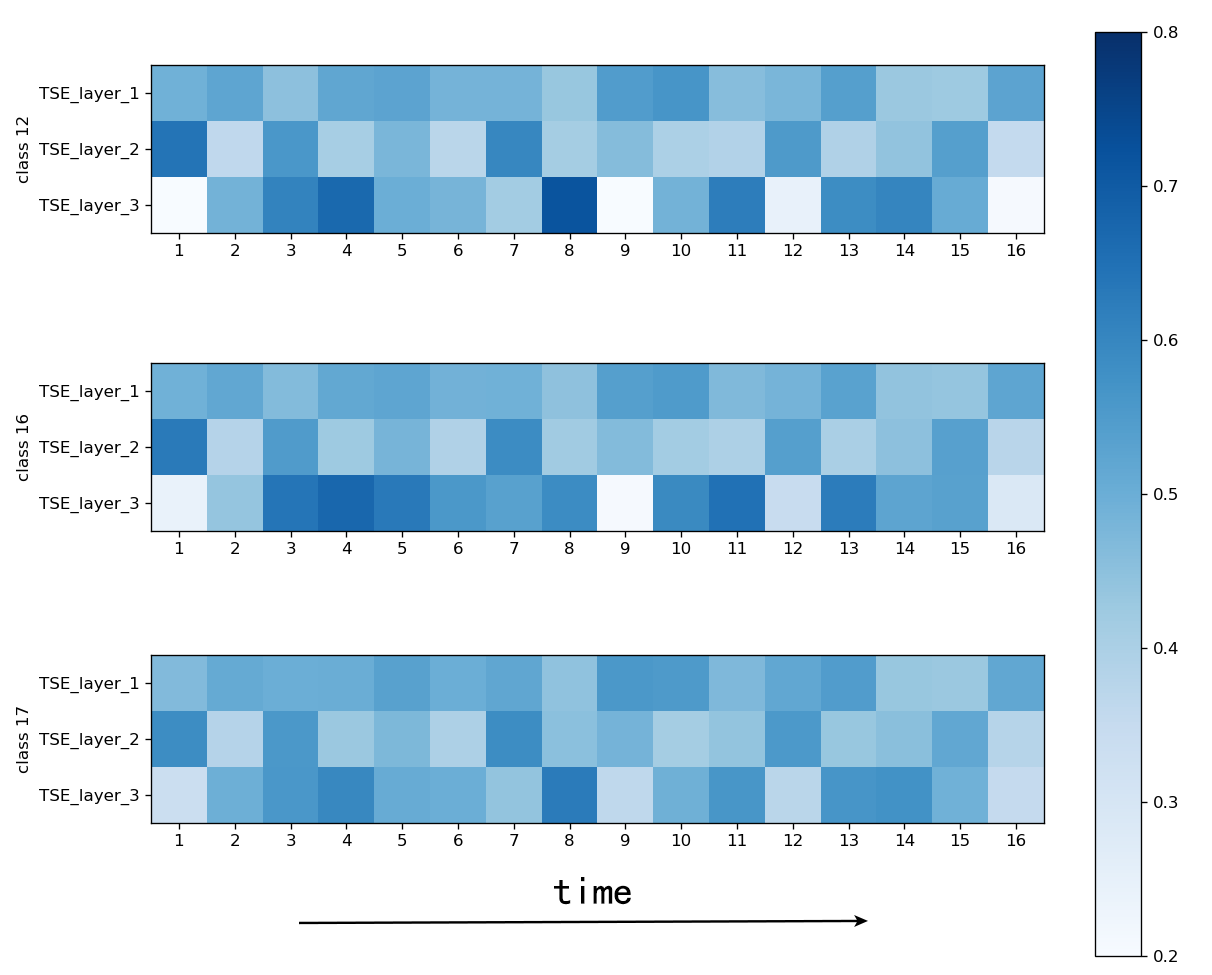}
\caption{Examples of temporal weights.}
\label{fig:example Excitation}
\end{center}
\end{figure}

To validate the effect of the proposed TSENet layers, the accuracy obtained by vanilla TCN is also shown in Table \ref{table:VIVA}. It can be seen that the presence of the TSENet layers in the TCN can improve the recognition rate by around 0.8\%. Three examples of the temporal weights produced by TSENet layers are shown in Figure 6. It is interesting to see that the weights in the third layer contain obvious large and small values, which means it does select the important ones from the short-term features. Moreover, if we change the 3D-Dense networks to Res3D which is used for extracting the short-term features. The accuracy will further drop about 4.8\%. It proves the effectiveness of the structure of the proposed model. 

\begin{figure*} [tp]
    \centering
	  \subfloat[confusion matrix on the VIVA dataset]{\label{fig:cm1a}
       \includegraphics[width=.48\linewidth]{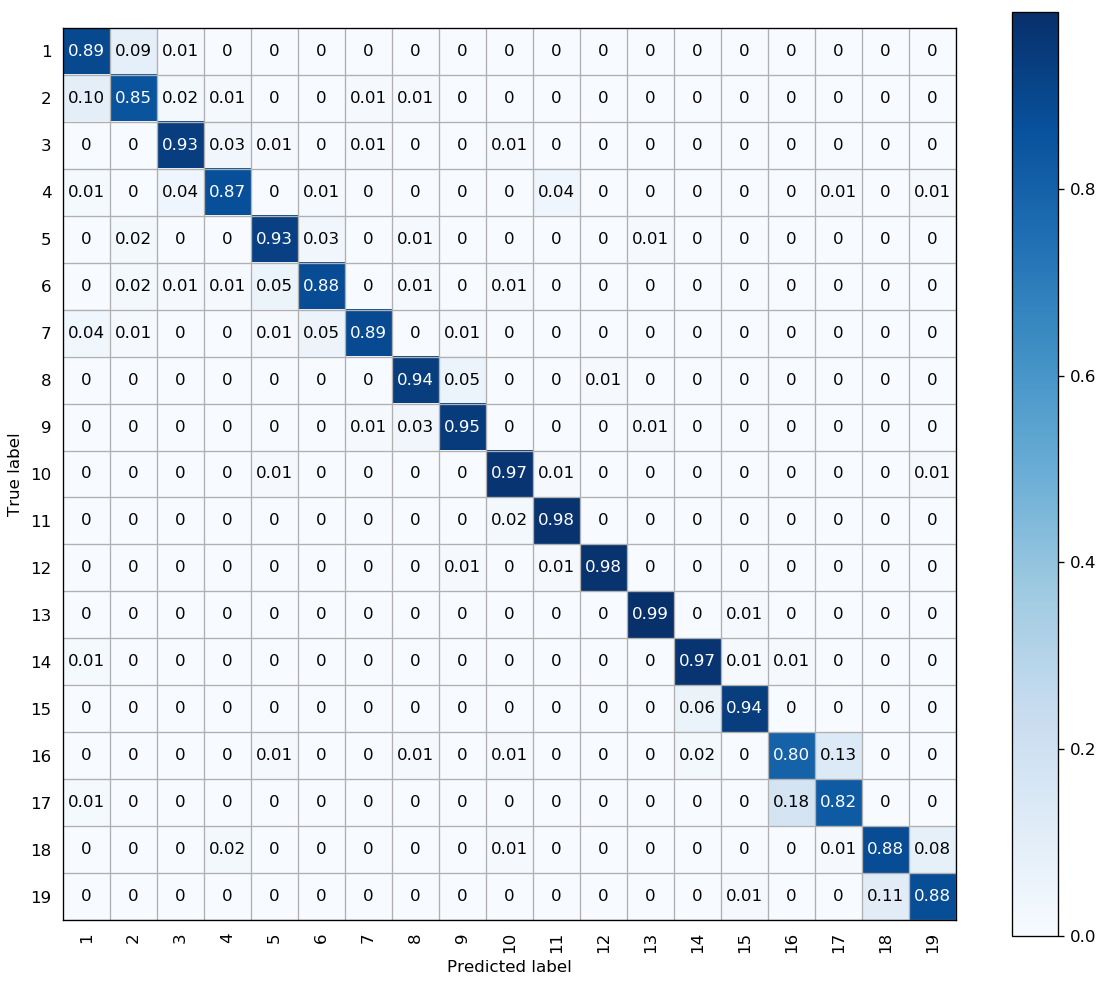}}
    \hfill
	  \subfloat[confusion matrix on the NVGesture dataset]{\label{fig:cm1b}
        \includegraphics[width=0.48\linewidth]{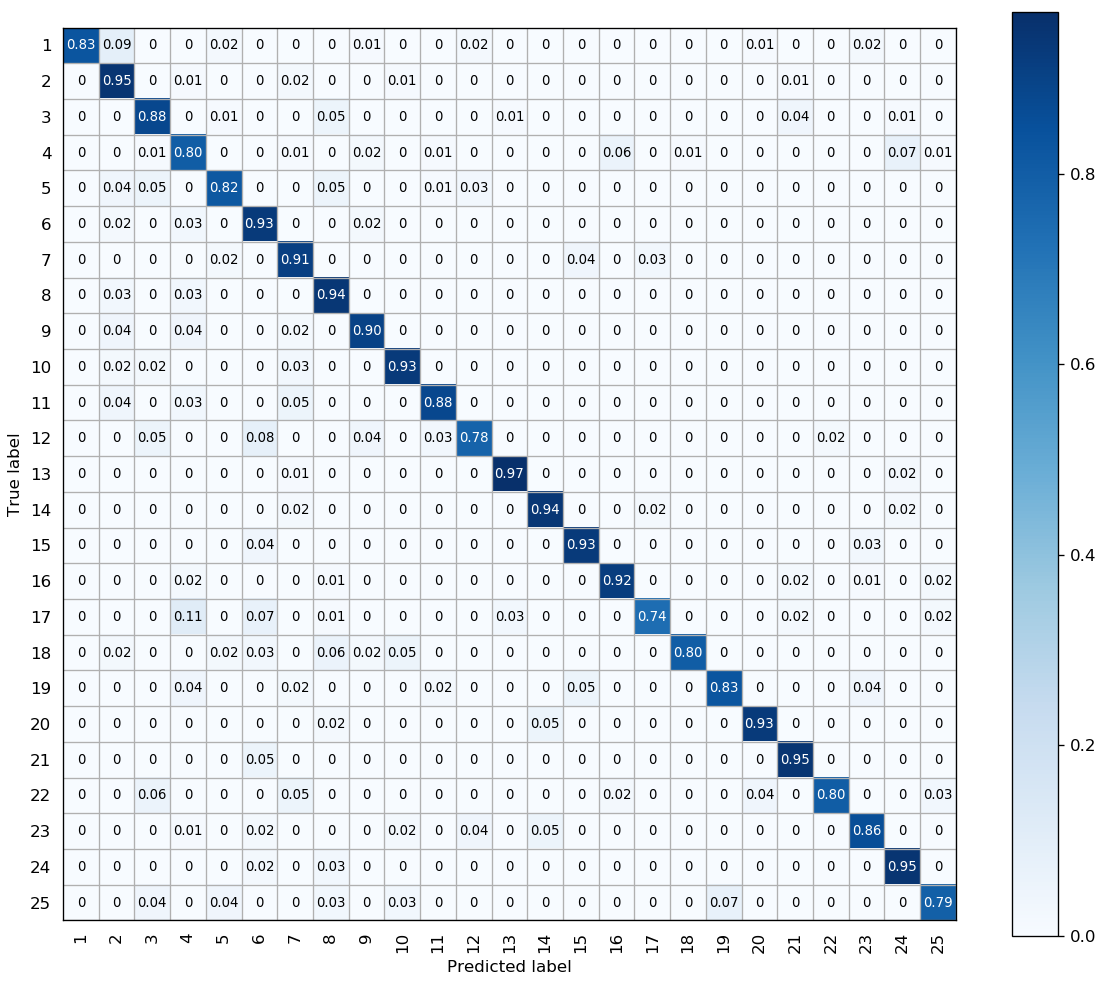}}
    \\
	  \caption{The confusion matrices obtained by comparing the grand-truth labels and the predicted labels from the RGB+depth modalities on the VIVA dataset and the RGB+opt.flow+depth modalities on the NVGesture dataset by our model. Best seen on the computer, in color and zoomed in.} 
	  \label{fig1}
\end{figure*}

Figure \ref{fig:cm1a} shows the confusion matrix as well for the experiment. It can be seen that the proposed model confused between the Swipe and Scroll gestures performed along the same direction.  Many gestures were mis-classified as the Swipe down gesture, the Rotate CW/CCW gestures were difficult for the proposed model. In some case, the propose model may have difficulties with distinguishing between the Swipe + and the Swipe X gestures. 

\subsection{EVALUATION ON NVgesture.}

The NVGesture dataset, containing RGB, depth and optical flow modalities, is also used to test the proposed model. Table \ref{table:nvGesture} tabulates the results of our method in comparison with the recent state-of-the-art methods: HOG+HOG2\cite{ohn2014hand}, improved dense trajectories(iDT)\cite{wang2016robust}, R3DCNN\cite{molchanov2016online}, two-stream CNNs\cite{simonyan2014two}, and C3D as well as human labeling accuracy. 
The iDT method is often recognized as the best performing hand-crafted method. However, we observe that similar to the pervious experiments the 3D-CNN-based methods outperform the other hand gesture recognition methods, and among them, our method provides the better performance in all the modalities. 
Nonetheless, compare to the latest method MTUT, our method accuacies are close to the MTUT. 
Our method has the better performance in both of RGB and optical flow modalities, it improve accuracy by 0.73\%.
But in RGB+Depth modalities and in RGB+Depth+Opt.flow modalities, our method is not performing good enough.
This is in part due to the knowledge that gestures in NVGesture are more complex and have more invalid information. Although through TCN and TSE, our method can key information in the frames and weaken the influence of irrelevant information, the redundant non gesture information, especially in temporal, always affects the final results of the experiment.

\begin{table}[htbp]
\caption{\textbf{Accuracies of different multimodal fusion-based hand gesture methods on the NVGesture dataset. The top performer is denoted by boldface.}}
\label{table:nvGesture}
\centering
\setlength{\tabcolsep}{3pt}
\begin{tabular}{lcc}
\toprule
Method & Fused modalities &Accuracy\\  
  \midrule
HOG+HOG2\cite{ohn2014hand}&RGB+Depth&36.9 \\
I3D\cite{carreira2017quo}     &RGB+Depth&83.82 \\
MTUT\cite{abavisani2019improving} &RGB+Depth&\textbf{86.10}\\
Ours    &RGB+Depth&84.87 \\
  \midrule
Two Stream CNNs\cite{simonyan2014two}&RGB+Opt. flow&65.6 \\
iDT\cite{wang2016robust}&RGB+Opt. flow& 73.4\\
I3D\cite{carreira2017quo} &RGB+Opt. flow&84.43\\
MTUT\cite{abavisani2019improving} &RGB+Opt. flow&85.48 \\
Ours&RGB+Opt. flow&\textbf{86.21}\\

  \midrule
R3DCNN\cite{molchanov2016online}&RGB+Depth+Opt. flow&83.8 \\
I3D\cite{carreira2017quo}     &RGB+Depth+Opt. flow&85.68 \\
MTUT\cite{abavisani2019improving}    &RGB+Depth+Opt. flow&\textbf{86.93}\\
Ours&RGB+Depth+Opt. flow&86.37\\
  \midrule
  \midrule
\multicolumn{2}{l}{Human labeling accuracy:}&88.4\\
  \bottomrule

\end{tabular}
\label{tab1}
\end{table}
\section{Conclusion}
\label{sec:conclusion}

We developed an effective method for multi-modal (RGB, depth and optic flow data) dynamic hand gesture recognition with 3D-DenseNets and TCNs. And in TCNs, we improved and applied an attention model named SENets to learn and extract deeper temporal features. The experiments show that the proposed model achieved the highest accuracy in VIVA dataset, as well as competitive results  in NVGesture dataset.

However, our model is still not an end-to-end model and has to be trained step by step. Meanwhile, NVGesture still have a large room for improvement, we still have a lot of work to enhance the accuracy of the model.




%
%
%
\bibliographystyle{IEEEtran}
\normalem
\bibliography{reff}

\end{document}